\documentclass[10pt,journal,compsoc]{IEEEtran}

\usepackage{times}
\usepackage{epsfig}
\usepackage{graphicx}
\usepackage{color}
\usepackage{amsmath}
\usepackage{amssymb}
\usepackage[ruled,vlined,boxed]{algorithm2e}
\usepackage{multirow}
\usepackage[table]{xcolor} 

\definecolor{LightCyan}{rgb}{0.88,1,1}

\ifCLASSOPTIONcompsoc
  \usepackage[nocompress]{cite}
\else
  \usepackage{cite}
\fi
\ifCLASSINFOpdf
\else
\fi
\usepackage{hyperref}

\begin{document}
%
\title{SegNet: A Deep Convolutional Encoder-Decoder Architecture for Image Segmentation}

\author{Vijay Badrinarayanan, Alex Kendall, Roberto Cipolla,~\IEEEmembership{Senior Member,~IEEE,}       

\IEEEcompsocitemizethanks{\IEEEcompsocthanksitem V. Badrinarayanan, A. Kendall, R. Cipolla are with the Machine Intelligence Lab, Department of Engineering,
University of Cambridge, UK. 
\protect\\
E-mail: vb292,agk34,cipolla@eng.cam.ac.uk
} }
\IEEEtitleabstractindextext{%
\begin{abstract}
We present a novel and practical deep fully convolutional neural network architecture for semantic pixel-wise segmentation termed SegNet. This core trainable segmentation engine consists of an  encoder network, a corresponding decoder network followed by a pixel-wise classification layer. The architecture of the encoder network is topologically identical to the 13 convolutional layers in the VGG16 network \cite{simonyan2014very}. The role of the decoder network is to map the low resolution encoder feature maps to full input resolution feature maps for pixel-wise classification. The novelty of SegNet lies is in the manner in which the decoder upsamples its lower resolution input feature map(s). Specifically, the decoder uses pooling indices computed in the max-pooling step of the corresponding encoder to perform non-linear upsampling. This eliminates the need for learning to upsample. The upsampled maps are sparse and are then convolved with trainable filters to produce dense feature maps. We compare our proposed architecture with the widely adopted FCN \cite{FCN} and also with the well known DeepLab-LargeFOV \cite{liang2015semantic}, DeconvNet \cite{noh2015learning} architectures. This comparison reveals the memory versus accuracy trade-off involved in achieving good segmentation performance. 

\indent SegNet was primarily motivated by scene understanding applications. Hence, it is designed to be efficient both in terms of memory and computational time during inference. It is also significantly smaller in the number of trainable parameters than other competing architectures and can be trained end-to-end using stochastic gradient descent. We also performed a controlled benchmark of SegNet and other architectures on both road scenes and  SUN RGB-D indoor scene segmentation tasks. These quantitative assessments show that SegNet provides good performance with competitive inference time and most efficient inference memory-wise as compared to other architectures. We also provide a Caffe implementation of SegNet and a web demo at \url{http://mi.eng.cam.ac.uk/projects/segnet/}.  
\end{abstract}

}

\maketitle


%

\section{Introduction}
\label{sec:introduction}
\begin{figure*}[!t]
\center
\includegraphics[width=0.95\textwidth]{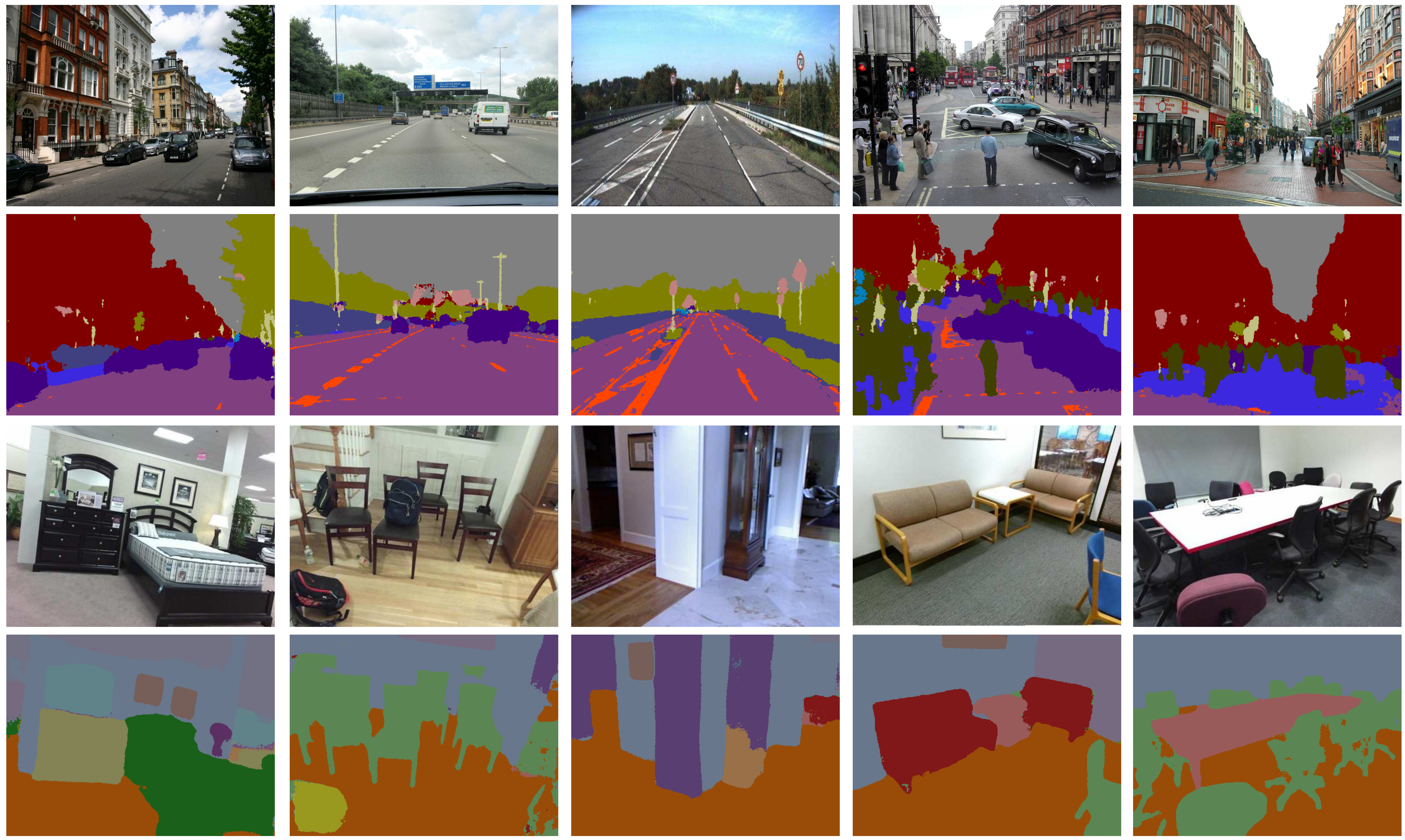}
\caption{\footnotesize{SegNet predictions on road scenes and indoor scenes. To try our system yourself, please see our online web demo at \url{http://mi.eng.cam.ac.uk/projects/segnet/}. }}
\label{Teaser}
\end{figure*} 

Semantic segmentation has a wide array of applications ranging from scene understanding, inferring support-relationships among objects to autonomous driving. Early methods that relied on low-level vision cues have fast been superseded by popular machine learning algorithms. In particular, deep learning has seen huge success lately in handwritten digit recognition, speech, categorising whole images and detecting objects in images \cite{szegedy2015going, Simonyan}. Now there is an active interest for semantic pixel-wise labelling \cite{FarabetPAMI} \cite{raey},  \cite{socher2011parsing},\cite{FCN},\cite{noh2015learning},\cite{zheng2015conditional},\cite{ParseNetRabinovich},\cite{SegNetarXiv},\cite{eigen2015predicting}, \cite{liang2015semantic}, \cite{papandreou2015weakly},\cite{yu2015multi}, \cite{ronneberger2015u}. 
However, some of these recent approaches have tried to directly adopt deep architectures designed for category prediction to pixel-wise labelling \cite{FarabetPAMI}. The results, although very encouraging, appear coarse \cite{liang2015semantic}. This is primarily because max pooling and sub-sampling reduce feature map resolution. Our motivation to design SegNet arises from this need to map low resolution features to input resolution for pixel-wise classification. This mapping must produce features which are useful for accurate boundary localization. 

Our architecture, SegNet, is designed to be an efficient architecture  for pixel-wise semantic segmentation. It is primarily motivated by road scene understanding applications which require the ability to model appearance (road, building), shape (cars, pedestrians) and understand the spatial-relationship (context) between different classes such as road and side-walk. In typical road scenes, the majority of the pixels belong to large classes such as road, building and hence the network must produce smooth segmentations. The engine must also have the ability to delineate objects based on their shape despite their small size. Hence it is important to retain boundary information in the extracted image representation. From a computational perspective, it is necessary for the network to be efficient in terms of both memory and computation time during inference. The ability to train end-to-end in order to jointly optimise all the weights in the network using an efficient weight update technique such as stochastic gradient descent (SGD) \cite{Bottou} is an additional benefit since it is more easily repeatable. The design of SegNet arose from a need to match these criteria.

The encoder network in SegNet is topologically identical to the convolutional layers in VGG16 \cite{simonyan2014very}. We remove the fully connected layers of VGG16 which makes the SegNet encoder network significantly smaller and easier to train than many other recent architectures \cite{FCN,noh2015learning,ParseNetRabinovich,hong2015decoupled}. The key component of SegNet is the decoder network which consists of a hierarchy of decoders one corresponding to each encoder. Of these, the appropriate decoders use the max-pooling indices received from the corresponding encoder to perform non-linear upsampling of their input feature maps. This idea was inspired from an architecture designed for unsupervised feature learning \cite{Ranzato}. Reusing max-pooling indices in the decoding process has several practical advantages; (i) it improves boundary delineation , (ii) it reduces the number of  parameters enabling end-to-end training, and (iii) this form of upsampling can be incorporated into any encoder-decoder architecture such as \cite{FCN,zheng2015conditional} with only a little modification. 

One of the main contributions of this paper is our analysis of the SegNet decoding technique and the widely used Fully Convolutional Network (FCN) \cite{FCN}. This is in order to convey the practical trade-offs involved in designing segmentation architectures. Most recent deep architectures for segmentation have identical encoder networks, i.e VGG16, but differ in the form of the decoder network, training and inference. Another common feature is they have trainable parameters in the order of hundreds of millions and thus encounter difficulties in performing end-to-end training \cite{noh2015learning}. The difficulty of training these networks has led to multi-stage training \cite{FCN}, appending networks to a pre-trained architecture such as FCN \cite{zheng2015conditional}, use of supporting aids such as region proposals for inference \cite{noh2015learning}, disjoint training of classification and segmentation networks \cite{hong2015decoupled} and use of additional training data for pre-training \cite{ParseNetRabinovich} \cite{mottaghi2014role} or for full training \cite{zheng2015conditional}. In addition, performance boosting post-processing techniques \cite{liang2015semantic} have also been popular. Although all these factors improve performance on challenging benchmarks \cite{Pascal}, it is unfortunately difficult from their quantitative results to disentangle the key design factors necessary to achieve good performance. We therefore analysed the decoding process used in some of these approaches \cite{FCN,noh2015learning} and reveal their pros and cons.

We evaluate the performance of SegNet on two scene segmentation tasks, CamVid road scene segmentation \cite{GabeDataset} and SUN RGB-D indoor scene segmentation \cite{song2015sun}. Pascal VOC12 \cite{Pascal} has been the benchmark challenge for segmentation over the years. However, the majority of this task has one or two foreground classes surrounded by a highly varied background. This implicitly favours techniques used for detection as shown by the recent work on a decoupled classification-segmentation network \cite{hong2015decoupled} where the classification network can be trained with a large set of weakly labelled data and the independent segmentation network performance is improved. The method of \cite{liang2015semantic} also use the feature maps of the classification network with an independent CRF post-processing technique to perform segmentation. The performance can also be boosted by the use additional inference aids such as region proposals \cite{noh2015learning}, \cite{zitnick2014edge}. Therefore, it is different from scene understanding where the idea is to exploit co-occurrences of objects and other spatial-context to perform robust segmentation. To demonstrate the efficacy of SegNet, we present a real-time online demo of road scene segmentation into 11 classes of interest for autonomous driving (see link in Fig. \ref{Teaser}). Some example test results produced on randomly sampled road scene images from Google and indoor test scenes from the SUN RGB-D dataset \cite{song2015sun} are shown in Fig. \ref{Teaser}. 

The remainder of the paper is organized as follows. In Sec. \ref{LitReview} we review related recent literature. We describe the SegNet architecture and its analysis in Sec. \ref{Architecture}. In Sec. \ref{Benchmarking} we evaluate the performance of SegNet on outdoor and indoor scene datasets. This is followed by a general discussion regarding our approach with pointers to future work in Sec. \ref{Discussion}. We conclude in Sec. \ref{Conclusion}. 

\section{Literature Review}
\label{LitReview}

Semantic pixel-wise segmentation is an active topic of research, fuelled by challenging datasets \cite{GabeDataset,silberman2012indoor,GeigerKITTI,Pascal,song2015sun}. Before the arrival of deep networks, the best performing methods mostly relied on hand engineered features classifying pixels independently. Typically, a patch is fed into a classifier \textit{e.g.} Random Forest \cite{Jamie2,Brostow} or Boosting \cite{Sturgess,LadickyECCV} to predict the class probabilities of the center pixel. Features based on appearance \cite{Jamie2} or SfM and appearance \cite{Brostow,Sturgess, LadickyECCV} have been explored for the CamVid road scene understanding test \cite{GabeDataset}. These per-pixel noisy predictions (often called \textit{unary} terms) from the classifiers are then smoothed by using a pair-wise or higher order CRF \cite{Sturgess,LadickyECCV} to improve the accuracy. More recent approaches have aimed to produce high quality unaries by trying to predict the labels for all the pixels in a patch as opposed to only the center pixel. This improves the results of Random Forest based unaries \cite{kontschieder2011structured} but thin structured classes are classified poorly. Dense depth maps computed from the CamVid video have also been used as input for classification using Random Forests \cite{zhang2010semantic}. Another approach argues for the use of a combination of popular hand designed features and spatio-temporal super-pixelization to obtain higher accuracy \cite{tighe2013superparsing}. The best performing technique on the CamVid test \cite{LadickyECCV} addresses the imbalance among label frequencies by combining object detection outputs with classifier predictions in a CRF framework. The result of all these techniques indicate the need for improved features for  classification.  

Indoor RGBD pixel-wise semantic segmentation has also gained popularity since the release of the NYU dataset \cite{silberman2012indoor}. This dataset showed the usefulness of the depth channel to improve segmentation. Their approach used features such as RGB-SIFT, depth-SIFT and pixel location as input to a neural network classifier to predict pixel unaries. The noisy unaries are then smoothed using a CRF. Improvements were made using a richer feature set including LBP and region segmentation to obtain higher accuracy \cite{ren2012rgb} followed by a CRF. In more recent work \cite{silberman2012indoor}, both class segmentation and support relationships are inferred together using a combination of RGB and depth based cues. Another approach focuses on real-time joint reconstruction and semantic segmentation, where Random Forests are used as the classifier \cite{Hermans14ICRA}. Gupta \emph{et~al.}  \cite{gupta2013perceptual} use boundary detection and hierarchical grouping before performing category segmentation. The common attribute in all these approaches is the use of hand engineered features for classification of either RGB or RGBD images. 

The success of deep convolutional neural networks for object classification has more recently led researchers to exploit their feature learning capabilities for structured prediction problems such as segmentation. There have also been attempts to apply networks designed for object categorization to segmentation, particularly by replicating the deepest layer features in blocks to match image dimensions \cite{FarabetPAMI,FarabetPurityCover,Grangier,Gatta}. However, the resulting classification is blocky \cite{Grangier}. Another approach using recurrent neural networks \cite{pinheiro2014recurrent} merges several low resolution predictions to create input image resolution predictions. These techniques are already an improvement over hand engineered features \cite{FarabetPAMI} but their ability to delineate boundaries is poor. 

Newer deep architectures \cite{FCN,noh2015learning,eigen2015predicting,hong2015decoupled,zheng2015conditional} particularly designed for segmentation have advanced the state-of-the-art by learning to decode or map low resolution image representations to pixel-wise predictions. The encoder network which produces these low resolution representations in all of these architectures is the VGG16 classification network \cite{simonyan2014very} which has $13$ convolutional layers and $3$ fully connected layers. This encoder network weights are typically pre-trained on the large ImageNet object classification dataset \cite{ImageNet}. The decoder network varies between these architectures and is the part which is responsible for producing multi-dimensional features for each pixel for classification.

Each decoder in the Fully Convolutional Network (FCN) architecture \cite{FCN} learns to upsample its input feature map(s) and combines them with the corresponding encoder feature map to produce the input to the next decoder. It is an architecture which has a large number of trainable parameters in the encoder network (134M) but a very small decoder network (0.5M). The overall large size of this network makes it hard to train end-to-end on a relevant task. Therefore, the authors use a stage-wise training process. Here each decoder in the decoder network is progressively added to an existing trained network. The network is grown until no further increase in performance is observed. This growth is stopped after three decoders thus ignoring high resolution feature maps can certainly lead to loss of edge information \cite{noh2015learning}. Apart from training related issues, the need to reuse the encoder feature maps in the decoder makes it memory intensive in test time. We study this network in more detail as it the core of other recent architectures \cite{ParseNetRabinovich, zheng2015conditional}.

\begin{figure*}[!ht]
\center
\includegraphics[width=0.95\textwidth]{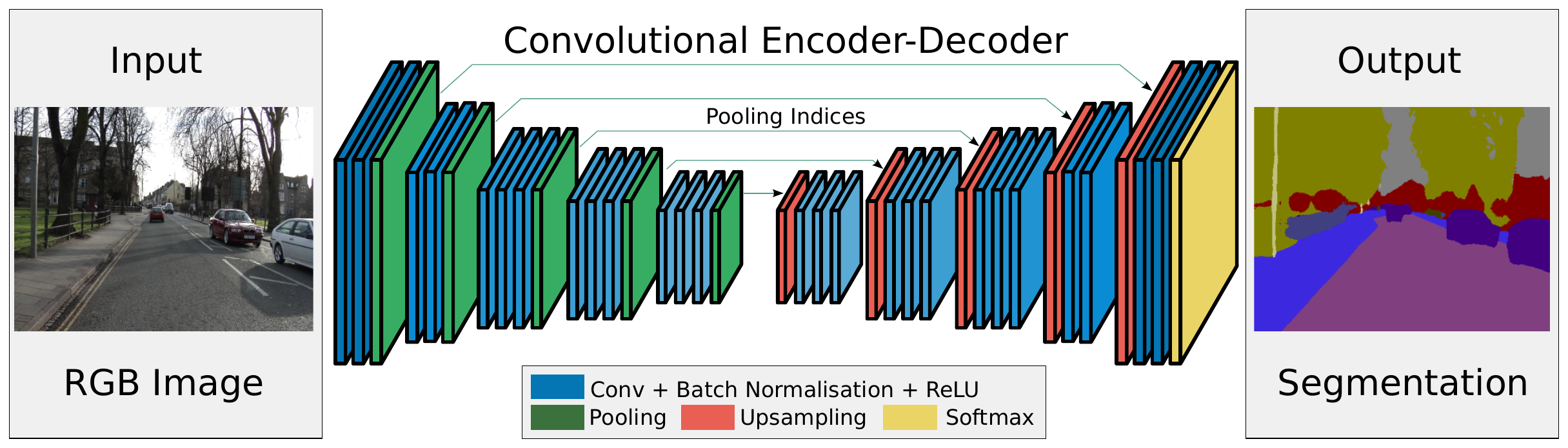}
\caption{\footnotesize{An illustration of the SegNet architecture. There are no fully connected layers and hence it is only convolutional. A decoder upsamples its input using the transferred pool indices from its encoder to produce a sparse feature map(s). It then performs convolution with a trainable filter bank to densify the feature map. The final decoder output feature maps are fed to a soft-max classifier for pixel-wise classification.}}
\label{SegNetArchitecture}
\end{figure*}

The predictive performance of FCN has been improved further by appending the FCN with a recurrent neural network (RNN) \cite{zheng2015conditional} and fine-tuning them on large datasets \cite{Pascal},\cite{COCO}. The RNN layers mimic the sharp boundary delineation capabilities of CRFs while exploiting the feature representation power of FCN's. They show a significant improvement over FCN-8 but also show that this difference is reduced when more training data is used to train FCN-8. The main advantage of the CRF-RNN is revealed when it is jointly trained with an architecture such as the FCN-8. The fact that joint training helps is also shown in other recent results \cite{UrtasunSegmentation}, \cite{lin2015efficient}. Interestingly, the deconvolutional network \cite{noh2015learning} performs significantly better than FCN although at the cost of a more complex training and inference. This however raises the question as to whether the perceived advantage of the CRF-RNN would be reduced as the core feed-forward segmentation engine is made better. In any case, the CRF-RNN network can be appended to any deep segmentation architecture including SegNet.

Multi-scale deep architectures are also being pursued \cite{eigen2015predicting,lin2015efficient}. They come in two flavours, (i) those which use input images at a few scales and corresponding deep feature extraction networks, and (ii) those which combine feature maps from different layers of a single deep architecture \cite{hariharan2015hypercolumns} \cite{ParseNetRabinovich}. The common idea is to use features extracted at multiple scales to provide both local and global context \cite{mostajabi2015feedforward} and the using feature maps of the early encoding layers retain more high frequency detail leading to sharper class boundaries. Some of these architectures are difficult to train due to their parameter size \cite{eigen2015predicting}. Thus a multi-stage training process is employed along with data augmentation. The inference is also expensive with multiple convolutional pathways for feature extraction. Others \cite{lin2015efficient} append a CRF to their multi-scale network and jointly train them. However, these are not feed-forward at test time and require optimization to determine the MAP labels.

Several of the recently proposed deep architectures for segmentation are not feed-forward in inference time \cite{noh2015learning}, \cite{liang2015semantic}, \cite{hong2015decoupled}. They require either MAP inference over a CRF \cite{lin2015efficient}, \cite{UrtasunSegmentation} or aids such as region proposals \cite{noh2015learning} for inference. We believe the perceived performance increase obtained by using a CRF is due to the lack of good decoding techniques in their core feed-forward segmentation engine. SegNet on the other hand uses decoders to obtain features for accurate pixel-wise classification.

The recently proposed Deconvolutional Network \cite{noh2015learning} and its semi-supervised variant the Decoupled network \cite{hong2015decoupled} use the max locations of the encoder feature maps (pooling indices) to perform non-linear upsampling in the decoder network. The authors of these architectures, independently of SegNet (first submitted to CVPR 2015 \cite{SegNetarXiv}), proposed this idea of decoding in the decoder network. However, their encoder network consists of the fully connected layers from the VGG-16 network which consists of about $90\%$ of the parameters of their entire network. This makes training of their network very difficult and thus require additional aids such as the use of region proposals to enable training. Moreover, during inference these proposals are used and this increases inference time  significantly. From a benchmarking point of view, this also makes it difficult to evaluate the performance of their architecture (encoder-decoder network) without other aids. In this work we discard the fully connected layers of the VGG16 encoder network which enables us to train the network using the relevant training set using SGD optimization. Another recent method \cite{liang2015semantic} shows the benefit of reducing the number of parameters significantly without sacrificing performance, reducing memory consumption and improving inference time. 

Our work was inspired by the unsupervised feature learning architecture proposed by Ranzato \emph{et al.} \cite{Ranzato}. The key learning module is an encoder-decoder network. An encoder consists of convolution with a filter bank, element-wise tanh non-linearity, max-pooling and sub-sampling to obtain the feature maps. For each sample, the indices of the max locations computed during pooling are stored and passed to the decoder. The decoder upsamples the feature maps by using the stored pooled indices. It convolves this upsampled map using a trainable decoder filter bank to reconstruct the input image. This architecture was used for unsupervised pre-training for classification. A somewhat similar decoding technique is used for visualizing trained convolutional networks\cite{zeiler2010deconvolutional} for classification. The architecture of Ranzato \emph{et al.} mainly focused on layer-wise feature learning using small input patches. This was extended by Kavukcuoglu et. al. \cite{KorayUnsup} to accept full image sizes as input to learn hierarchical encoders. Both these approaches however did not attempt to use \textit{deep encoder-decoder} networks for unsupervised feature training as they discarded the decoders after each encoder training. Here, SegNet differs from these architectures as the deep encoder-decoder network is trained jointly for a supervised learning task and hence the decoders are an integral part of the network in test time.

Other applications where pixel wise predictions are made using deep networks are image super-resolution \cite{dong2014learning} and depth map prediction from a single image \cite{eigen2014NIPS}. The authors in \cite{eigen2014NIPS} discuss the need for learning to upsample from low resolution feature maps which is the central topic of this paper.  

\section{Architecture}
\label{Architecture}
SegNet has an encoder network and a corresponding decoder network, followed by a final pixelwise classification layer. This architecture is illustrated in Fig. \ref{Architecture}. The encoder network consists of $13$ convolutional layers which correspond to the first $13$ convolutional layers in the VGG16 network \cite{simonyan2014very} designed for object classification. We can therefore initialize the training process from weights trained for classification on large datasets \cite{ImageNet}. We can also discard the fully connected layers in favour of retaining higher resolution feature maps at the deepest encoder output. This also reduces the number of parameters in the SegNet encoder network  significantly (from 134M to 14.7M) as compared to other recent architectures \cite{FCN}, \cite{noh2015learning} (see. Table \ref{TimeBenchmark}). Each encoder layer has a corresponding decoder layer and hence the decoder network has $13$ layers. The final decoder output is fed to a multi-class soft-max classifier to produce class probabilities for each pixel independently. 

Each \textit{encoder} in the encoder network performs convolution with a filter bank to produce a set of feature maps. These are then batch normalized \cite{BN},\cite{badrinarayananunderstanding}). Then an element-wise rectified-linear non-linearity (ReLU) $max(0,x)$ is applied. Following that, max-pooling with a $2\times2$ window and stride $2$ (non-overlapping window) is performed and the resulting output is sub-sampled by a factor of $2$. Max-pooling is used to achieve translation invariance over small spatial shifts in the input image. Sub-sampling results in a large input image context (spatial window) for each pixel in the feature map. While  several layers of max-pooling and sub-sampling can achieve more translation invariance for robust classification correspondingly there is a loss of spatial resolution of the feature maps. The increasingly lossy (boundary detail) image representation is not beneficial for segmentation where boundary delineation is vital. Therefore, it is necessary to \textit{capture and store} boundary information in the encoder feature maps before sub-sampling is performed. If memory during inference is not constrained, then all the encoder feature maps (after sub-sampling) can be stored. This is usually not the case in practical applications and hence we propose a more efficient way to store this information. It involves storing only the max-pooling \textit{indices}, i.e, the locations of the maximum feature value in each pooling window is memorized for each encoder feature map. In principle, this can be done using 2 bits for each $2\times2$ pooling window and is thus much more efficient to store as compared to memorizing feature map(s) in float precision. As we show later in this work, this lower memory storage results in a slight loss of accuracy but is still suitable for practical applications.

\begin{figure*}
\centering
\includegraphics[width=0.6\textwidth]{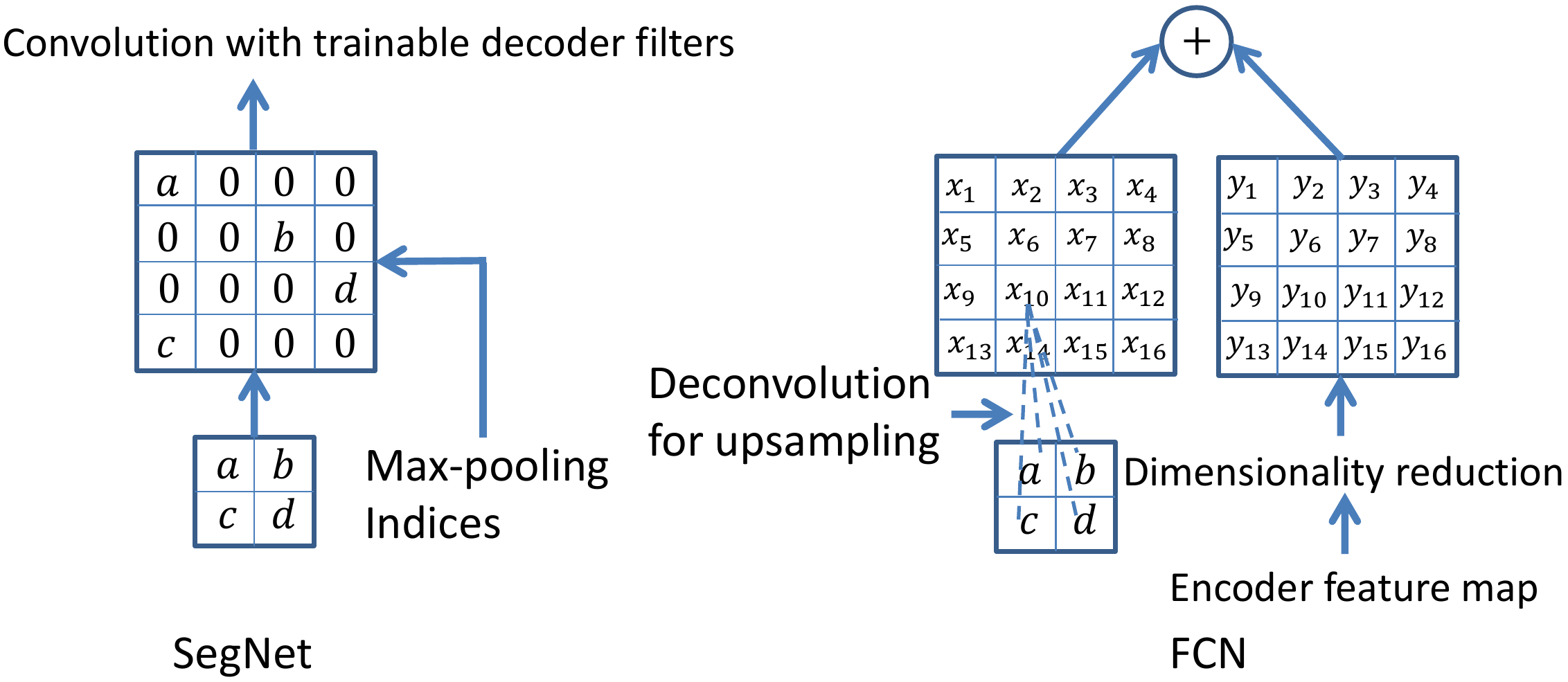}
\caption{\footnotesize{An illustration of SegNet and FCN \cite{FCN} decoders. $a,b,c,d$ correspond to values in a feature map. SegNet uses the max pooling indices to upsample (without learning) the feature map(s) and convolves with a trainable decoder filter bank. FCN upsamples by learning to deconvolve the input feature map and adds the corresponding encoder feature map to produce the decoder output. This feature map is the output of the max-pooling layer (includes sub-sampling) in the corresponding encoder. Note that there are no trainable decoder filters in FCN.}}
\label{Upsampling}
\end{figure*}

The appropriate \textit{decoder} in the decoder network upsamples its input feature map(s) using the memorized max-pooling indices from the corresponding encoder feature map(s). This step produces sparse feature map(s). This SegNet decoding technique is illustrated in Fig. \ref{Upsampling}. These feature maps are then convolved with a trainable decoder filter bank to produce dense feature maps. A batch normalization step is then applied to each of these maps. Note that the decoder corresponding to the first encoder (closest to the input image) produces a multi-channel feature map, although its encoder input has 3 channels (RGB). This is unlike the other decoders in the network which produce feature maps with the same number of size and channels as their encoder inputs. The high dimensional feature representation at the output of the final decoder is fed to a trainable soft-max classifier. This soft-max classifies each pixel independently. The output of the soft-max classifier is a K channel image of probabilities where K is the number of classes. The predicted segmentation corresponds to the class with maximum probability at each pixel.

We add here that two other architectures, DeconvNet \cite{NohDeconvNets} and U-Net \cite{ronneberger2015u} share a similar architecture to SegNet but with some differences. DeconvNet has a much larger parameterization, needs more computational resources and is harder to train end-to-end (Table \ref{TimeBenchmark}), primarily due to the use of fully connected layers (albeit in a convolutional manner) We report several comparisons with DeconvNet later in the paper Sec. \ref{Benchmarking}. 

As compared to SegNet, U-Net \cite{ronneberger2015u} (proposed for the medical imaging community) does not reuse pooling indices but instead transfers the entire feature map (at the cost of more memory) to the corresponding decoders and concatenates them to upsampled (via deconvolution) decoder feature maps. There is no conv5 and max-pool 5 block in U-Net as in the VGG net architecture. SegNet, on the other hand, uses all of the pre-trained convolutional layer weights from VGG net as pre-trained weights.

\begin{table*}[t]
\footnotesize
\centering
\tabcolsep=1pt
\begin{tabular}{c|c|c|c|cccc|ccc|cccc|ccc}
\multicolumn{4}{c}{} &  \multicolumn{7}{c}{Median frequency balancing} & \multicolumn{7}{|c}{Natural frequency balancing} \\
\hline
& & Storage & Infer & \multicolumn{4}{c|}{Test } &  \multicolumn{3}{c|}{Train } &   \multicolumn{4}{c|}{Test } & \multicolumn{3}{c}{Train }\\
Variant                   &Params (M) &  multiplier   &  time (ms)  & G       & C       & mIoU   &BF     &  G & C & mIoU   & G & C & mIoU  & BF & G & C & \multicolumn{1}{c}{mIoU}               \\ \hline \hline
\multicolumn{16}{c}{Fixed upsampling} \\ \hline
Bilinear-Interpolation           & 0.625 & 0 & 24.2 & 77.9 & 61.1 & 43.3 &20.83 & 89.1 & 90.2 &  82.7     & 82.7 & 52.5 & 43.8 &23.08 &93.5&74.1 &         59.9 \\ \hline                                                                                                   
\multicolumn{16}{c}{Upsampling using max-pooling indices}    \\                                                                                              
 \hline
SegNet-Basic                    & 1.425 & 1 & 52.6 & 82.7       &   62.0      &   47.7  &35.78  & 94.7 & 96. 2 & 92.7 &  84.0 & 54.6 & 46.3 & 36.67 & 96.1 & 83.9 &73.3                   \\ \hline
SegNet-Basic-EncoderAddition    &1.425 & 64  & 53.0 & 83.4  &     \textbf{63.6 }   &      48.5    &35.92    & 94.3 &  95.8 &  92.0 & \textbf{84.2} & 56.5 & \textbf{47.7} & 36.27 & 95.3&80.9 & 68.9\\ \hline
SegNet-Basic-SingleChannelDecoder& 0.625 &  1 & 33.1 &  81.2     &    60.7     &     46.1   & 31.62     & 93.2 & 94.8 &   90.3  & 83.5 & 53.9 & 45.2 & 32.45 &92.6 & 68.4 & 52.8        \\ \hline
\multicolumn{16}{c}{Learning to upsample (bilinear initialisation)}                                                                                                         
 \\ \hline
FCN-Basic                     & 0.65 & 11  & 24.2  &   81.7    &     62.4    &    47.3  &\textbf{38.11}      & 92.8 & 93.6 &  88.1  & 83.9 & 55.6 & 45.0 & \textbf{37.33} & 92.0 & 66.8 & 50.7       \\ \hline
FCN-Basic-NoAddition           &0.65 & n/a & 23.8 &   80.5    &    58.6     &     44.1    &31.96     & 92.5 & 93.0 &    87.2 & 82.3 & 53.9 & 44.2 & 29.43 &93.1& 72.8 & 57.6         \\ \hline
FCN-Basic-NoDimReduction                    &1.625 & 64 & 44.8  &  \textbf{84.1}     &  63.4       &   \textbf{50.1}    &37.37      & \textbf{95.1} & \textbf{96.5}  &  \textbf{93.2} &        83.5 & \textbf{57.3} & 47.0 & 37.13 &\textbf{97.2} & \textbf{91.7} & \textbf{84.8}
\\ \hline
FCN-Basic-NoAddition-NoDimReduction                    &1.625 & 0 & 43.9   & 80.5      &   61.6        & 45.9 & 30.47 &  92.5 & 94.6 & 89.9 &  83.7 & 54.8 & 45.5 & 33.17 & 95.0 & 80.2 & 67.8      \\ \hline 
\end{tabular}
\vspace*{0.1cm}
\caption{\footnotesize{Comparison of decoder variants. We quantify the performance using global (G), class average (C), mean of intersection over union (mIoU) and a semantic contour measure (BF). The testing and training accuracies are shown as percentages for both natural frequency and median frequency balanced training loss function. SegNet-Basic performs at the same level as FCN-Basic but requires only storing max-pooling indices and is therefore more memory efficient during inference. Note that the theoretical memory requirement reported is based only on the size of the first layer encoder feature map. FCN-Basic, SegNet-Basic, SegNet-Basic-EncoderAddition all have high BF scores indicating the need to use information in encoder feature maps for better class contour delineation. Networks with larger decoders and those using the encoder feature maps in full perform best, although they are least efficient in terms of inference time and memory.}}
\label{PoolingQuantCAFFE}
\end{table*}

\subsection{Decoder Variants}
\label{Variants}
Many segmentation architectures \cite{FCN, noh2015learning, liang2015semantic} share the same encoder network and they only vary in the form of their decoder network. 
Of these we choose to compare the SegNet decoding technique with the widely used Fully Convolutional Network (FCN) decoding technique \cite{FCN, zheng2015conditional}. 

In order to analyse SegNet and compare its performance with FCN (decoder variants) we use a smaller version of SegNet, termed \textbf{SegNet-Basic} \footnote{SegNet-Basic was earlier termed SegNet in a archival version of this paper \cite{SegNetarXiv}}, which has 4 encoders and 4 decoders. All the encoders in SegNet-Basic perform max-pooling and sub-sampling and the corresponding decoders upsample its input using the received max-pooling indices. Batch normalization is used after each convolutional layer in both the encoder and decoder network. No biases are used after convolutions and no ReLU non-linearity is present in the decoder network. Further, a constant kernel size of $7\times7$ over all the encoder and decoder layers is chosen to provide a wide context for smooth labelling \textit{i.e.} a pixel in the deepest layer feature map (layer $4$) can be traced back to a context window in the input image of $106\times106$ pixels. This small size of SegNet-Basic allows us to explore many different variants (decoders) and train them in reasonable time. Similarly we create \textbf{FCN-Basic}, a comparable version of FCN for our analysis which shares the same encoder network as SegNet-Basic but with the FCN decoding technique (see Fig. \ref{Upsampling}) used in all its decoders.  

On the left in Fig. \ref{Upsampling} is the decoding technique used by SegNet (also SegNet-Basic), where there is no learning involved in the upsampling step. However, the upsampled maps are convolved with trainable multi-channel decoder filters to densify its sparse inputs. Each decoder filter has the same number of channels as the number of upsampled feature maps. A smaller variant is one where the decoder filters are single channel, i.e they only convolve their corresponding upsampled feature map. This variant (\textbf{SegNet-Basic-SingleChannelDecoder}) reduces the number of trainable parameters and inference time significantly.

On the right in Fig. \ref{Upsampling} is the FCN (also FCN-Basic) decoding technique. The important design element of the FCN model is dimensionality reduction step of the encoder feature maps. This \textit{compresses} the encoder feature maps which are then used in the corresponding decoders. Dimensionality reduction of the encoder feature maps, say of 64 channels, is performed by convolving them with $1\times 1\times 64\times K$ trainable filters, where $K$ is the number of classes. The compressed $K$ channel final encoder layer feature maps are the input to the decoder network. In a decoder of this network, upsampling is performed by \textit{inverse convolution} using a fixed or trainable \textit{multi-channel upsampling kernel}. We set the kernel size to $8\times8$. This manner of upsampling is also termed as \textit{deconvolution}. Note that, in comparison, SegNet the multi-channel convolution using trainable decoder filters is performed after upsampling to densifying feature maps. The upsampled feature map in FCN has $K$ channels. It is then added element-wise to the corresponding resolution encoder feature map to produce the output decoder feature map. The upsampling kernels are initialized using bilinear interpolation weights \cite{FCN}. 

The FCN decoder model requires storing encoder feature maps during inference. This can be memory intensive for embedded applications; for e.g. storing 64 feature maps of the first layer of FCN-Basic at $180\times240$ resolution in 32 bit floating point precision takes 11MB. This can be made smaller using dimensionality reduction to the 11 feature maps which requires $\approx$ 1.9MB storage. SegNet on the other hand requires almost negligible storage cost for the pooling indices ($.17$MB if stored using 2 bits per $2\times2$ pooling window). We can also create a variant of the FCN-Basic model which discards the encoder feature map addition step and only learns the upsampling kernels (\textbf{FCN-Basic-NoAddition}). 

In addition to the above variants, we study upsampling using fixed bilinear interpolation weights which therefore requires no learning for upsampling (\textbf{Bilinear-Interpolation}). At the other extreme, we can add 64 encoder feature maps at each layer to the corresponding output feature maps from the SegNet decoder to create a more memory intensive variant of SegNet (\textbf{SegNet-Basic-EncoderAddition}).
Here both the pooling indices for upsampling are used, followed by a convolution step to densify its sparse input. This is then added element-wise to the corresponding encoder feature maps to produce a decoders output.

Another and more memory intensive FCN-Basic variant (\textbf{FCN-Basic-NoDimReduction}) is where there is no dimensionality reduction performed for the encoder feature maps. This implies that unlike FCN-Basic the final encoder feature map is not compressed to $K$ channels before passing it to the decoder network. Therefore, the number of channels at the end of each decoder is the same as the corresponding encoder (i.e $64$). 

We also tried other generic variants where feature maps are simply upsampled by \textit{replication} \cite{FarabetPAMI}, or by using a fixed (and sparse) array of indices for upsampling. These performed quite poorly in comparison to the above variants. A variant without max-pooling and sub-sampling in the encoder network (decoders are redundant) consumes more memory, takes longer to converge and performs poorly. 
Finally, please note that to encourage reproduction of our results we release the Caffe implementation of all the variants \footnote{See \url{ http://mi.eng.cam.ac.uk/projects/segnet/} for our SegNet code and web demo.}. 

\subsection{Training}
\label{Training}
We use the CamVid road scenes dataset to benchmark the performance of the decoder variants. This dataset is small, consisting of 367 training and 233 testing RGB images (day and dusk scenes) at $360\times480$ resolution. The challenge is to segment $11$ classes such as road, building, cars, pedestrians, signs, poles, side-walk etc. We perform local contrast normalization \cite{Jarrett} to the RGB input.

The encoder and decoder weights were all initialized using the technique described in He \emph{et~al.} \cite{he2015delvingICCV}. To train all the variants we use stochastic gradient descent (SGD) with a fixed learning rate of 0.1 and momentum of 0.9 \cite{Bottou} using our Caffe implementation of SegNet-Basic \cite{jia2014caffeACM}. We train the variants until the training loss converges. Before each epoch, the training set is shuffled and each mini-batch (12 images) is then picked in order thus ensuring that each image is used only once in an epoch. We select the model which performs highest on a validation dataset.

We use the cross-entropy loss \cite{FCN} as the objective function for training the network. The loss is summed up over all the pixels in a mini-batch. When there is large variation in the number of pixels in each class in the training set (e.g road, sky and building pixels dominate the CamVid dataset) then there is a need to weight the loss differently based on the true class. This is termed \textit{class balancing}. We use \textit{median frequency balancing } \cite{eigen2015predicting} where the weight assigned to a class in the loss function is the ratio of the median of class frequencies computed on the entire training set divided by the class frequency. This implies that larger classes in the training set have a weight smaller than $1$ and the weights of the smallest classes are the highest. We also experimented with training the different variants without class balancing or equivalently using \textit{natural frequency balancing}.

\subsection{Analysis}
\label{Analysis}
To compare the quantitative performance of the different decoder variants, we use three commonly used performance measures: global accuracy (G) which measures the percentage of pixels correctly classified in the dataset, class average accuracy (C) is the mean of the predictive accuracy over all classes and mean intersection over union (mIoU) over all classes as used in the Pascal VOC12 challenge \cite{Pascal}. The mIoU metric is a more stringent metric than class average accuracy since it penalizes false positive predictions. However, mIoU metric is not optimized for directly through the class balanced cross-entropy loss. 

The mIoU metric otherwise known as the Jacard Index is most commonly used in benchmarking. However,  Csurka et al. \cite{csurka2013good} note that this metric does not always correspond to human qualitative judgements (ranks) of good quality segmentation. They show with examples that mIoU favours region smoothness and does not evaluate boundary accuracy, a point also alluded to recently by the authors of FCN \cite{FCNnew}. Hence they propose to complement the mIoU metric with a boundary measure based on the Berkeley contour matching score commonly used to evaluate unsupervised image segmentation quality \cite{martin2004learning}. Csurka et al. \cite{csurka2013good} simply extend this to semantic segmentation and show that the measure of semantic contour accuracy used in conjunction with the mIoU metric agrees more with human ranking of segmentation outputs. 

The key idea in computing a semantic contour score is to evaluate the F1-measure \cite{martin2004learning} which involves computing the precision and recall values between the predicted and ground truth class boundary given a pixel tolerance distance. We used a value of $0.75\%$ of the image diagonal as the tolerance distance. The F1-measure for each class that is present in the ground truth test image is averaged to produce an image F1-measure. Then we compute the whole test set average, denoted the boundary F1-measure (BF) by average the image F1 measures.

We test each architectural variant after each $1000$ iterations of optimization on the CamVid validation set until the training loss converges. With a training mini-batch size of 12 this corresponds to testing approximately every 33 epochs (passes) through the training set. We select the iteration wherein the global accuracy is highest amongst the evaluations on the validation set. We report all the three measures of performance at this point on the held-out CamVid test set. Although we use class balancing while training the variants, it is still important to achieve high global accuracy to result in an overall smooth segmentation. Another reason is that the contribution of segmentation towards autonomous driving is mainly for delineating classes such as roads, buildings, side-walk, sky. These classes dominate the majority of the pixels in an image and a high global accuracy corresponds to good segmentation of these important classes. We also observed that reporting the numerical performance when class average is highest can often correspond to low global accuracy indicating a perceptually noisy segmentation output.

In Table \ref{PoolingQuantCAFFE} we report the numerical results of our analysis. We also show the size of the trainable parameters and the highest resolution feature map or pooling indices storage memory, i.e, of the first layer feature maps after max-pooling and sub-sampling. We show the average time for one forward pass with our Caffe implementation, averaged over $50$ measurements using a $360\times480$ input on an NVIDIA Titan GPU with cuDNN v3 acceleration. We note that the upsampling layers in the SegNet variants are not optimised using cuDNN acceleration. We show the results for both testing and training for all the variants at the selected iteration. The results are also tabulated without class balancing (natural frequency) for training and testing accuracies. Below we analyse the results with class balancing.

From the Table \ref{PoolingQuantCAFFE}, we see that bilinear interpolation based upsampling without any learning performs the worst based on all the measures of accuracy. All the other methods which either use learning for upsampling (FCN-Basic and variants) or learning decoder filters after upsampling (SegNet-Basic and its variants) perform significantly better. This emphasizes the need to learn decoders for segmentation. This is also supported by experimental evidence gathered by other authors when comparing FCN with SegNet-type decoding techniques\cite{noh2015learning}.

When we compare SegNet-Basic and FCN-Basic we see that both perform equally well on this test over all the measures of accuracy. The difference is that SegNet uses less \textbf{memory} during inference since it only stores max-pooling indices. On the other hand FCN-Basic stores \textbf{encoder feature maps} in full which consumes much more memory (11 times more). SegNet-Basic has a decoder with 64 feature maps in each decoder layer. In comparison FCN-Basic, which uses dimensionality reduction, has fewer (11) feature maps in each decoder layer. This reduces the number of convolutions in the decoder network and hence FCN-Basic is faster during inference (forward pass). From another perspective, the decoder network in SegNet-Basic makes it overall a larger network than FCN-Basic. This endows it with more flexibility and hence achieves higher training accuracy than FCN-Basic for the same number of iterations. Overall we see that SegNet-Basic has an advantage over FCN-Basic when inference time memory is constrained but where inference time can be compromised to some extent. 

SegNet-Basic is most similar to FCN-Basic-NoAddition in terms of their decoders, although the decoder of SegNet is larger. Both learn to produce dense feature maps, either directly by learning to perform deconvolution as in FCN-Basic-NoAddition or by first upsampling and then convolving with trained decoder filters. 
The performance of SegNet-Basic is superior, in part due to its larger decoder size. The accuracy of FCN-Basic-NoAddition is also lower as compared to FCN-Basic. This shows that it is vital to capture the information present in the encoder feature maps for better performance. In particular, note the large drop in the BF measure between these two variants. This can also explain the part of the reason why SegNet-Basic outperforms FCN-Basic-NoAddition. 

The size of the FCN-Basic-NoAddition-NoDimReduction model is slightly larger than SegNet-Basic since the final encoder feature maps are not compressed to match the number of classes $K$. This makes it a fair comparison in terms of the size of the model. The performance of this FCN variant is poorer than SegNet-Basic in test but also its training accuracy is lower for the same number of training epochs. This shows that using a larger decoder is not enough but it is also important to capture encoder feature map information to learn better, particular the fine grained contour information (notice the drop in the BF measure). Here it is also interesting to see that SegNet-Basic has a competitive training accuracy when compared to larger models such FCN-Basic-NoDimReduction. 

Another interesting comparison between FCN-Basic-NoAddition and SegNet-Basic-SingleChannelDecoder shows that using max-pooling indices for upsampling and an overall larger decoder leads to better performance. This also lends evidence to SegNet being a good architecture for segmentation, particularly when there is a need to find a compromise between storage cost, accuracy versus inference time. In the best case, when both memory and inference time is not constrained, larger models such as FCN-Basic-NoDimReduction and SegNet-EncoderAddition are both more accurate than the other variants. Particularly, discarding dimensionality reduction in the FCN-Basic model leads to the best performance amongst the FCN-Basic variants with a high BF score. This once again emphasizes the trade-off involved between memory and accuracy in segmentation architectures.

The last two columns of Table \ref{PoolingQuantCAFFE} show the result when no class balancing is used (natural frequency). Here, we can observe that without weighting the results are poorer for all the variants, particularly for class average accuracy and mIoU metric. The global accuracy is the highest without weighting since the majority of the scene is dominated by sky, road and building pixels. Apart from this all the inference from the comparative analysis of variants holds true for natural frequency balancing too, including the trends for the BF measure. SegNet-Basic performs as well as FCN-Basic and is better than the larger FCN-Basic-NoAddition-NoDimReduction. The bigger but less efficient models FCN-Basic-NoDimReduction and SegNet-EncoderAddition perform better than the other variants.

We can now summarize the above analysis with the following general points.
\begin{enumerate}
\item The best performance is achieved when encoder feature maps are stored in full. This is reflected in the semantic contour delineation metric (BF) most clearly.
\item When memory during inference is constrained, then compressed forms of encoder feature maps (dimensionality reduction, max-pooling indices) can be stored and used with an appropriate decoder (e.g. SegNet type) to improve performance.
\item Larger decoders increase performance for a given encoder network.
\end{enumerate}

\section{Benchmarking}
\label{Benchmarking}
We quantify the performance of SegNet on two scene segmentation  benchmarks using our Caffe implementation \footnote{\label{webdemo} Our web demo and Caffe implementation is available for evaluation at \url{http://mi.eng.cam.ac.uk/projects/segnet/}}. The first task is road scene segmentation which is of current practical interest for various autonomous driving related problems. The second task is indoor scene segmentation which is of immediate interest to several augmented reality (AR) applications. The input RGB images for both tasks were $360\times480$. 

We benchmarked SegNet against several other well adopted deep architectures for segmentation such as FCN \cite{FCN}, DeepLab-LargFOV \cite{liang2015semantic} and DeconvNet \cite{noh2015learning}. Our objective was to understand the performance of these architectures when trained end-to-end on the same datasets. To enable end-to-end training we added batch normalization \cite{BN} layers after each convolutional layer. For DeepLab-LargeFOV, we changed the max pooling 3 stride to 1 to achieve a final predictive resolution of $45\times60$. We restricted the feature size in the fully connnected layers of DeconvNet to $1024$ so as to enable training with the same batch size as other models. Here note that the authors of DeepLab-LargeFOV \cite{liang2015semantic} have also reported little loss in performance by reducing the size of the fully connected layers.

 In order to perform a controlled benchmark we used the same SGD solver \cite{Bottou} with a fixed learning rate of $10^{-3}$ and momentum of $0.9$. The optimization was performed for more than 100 epochs through the dataset until no further performance increase was observed. Dropout of $0.5$ was added to the end of deeper convolutional layers in all models to prevent overfitting (see \url{http://mi.eng.cam.ac.uk/projects/segnet/tutorial.html} for example caffe prototxt). For the road scenes which have $11$ classes we used a mini-batch size of $5$ and for indoor scenes with $37$ classes we used a mini-batch size of $4$.



\begin{figure*}
	\centering
	\includegraphics[width=0.8\textwidth]{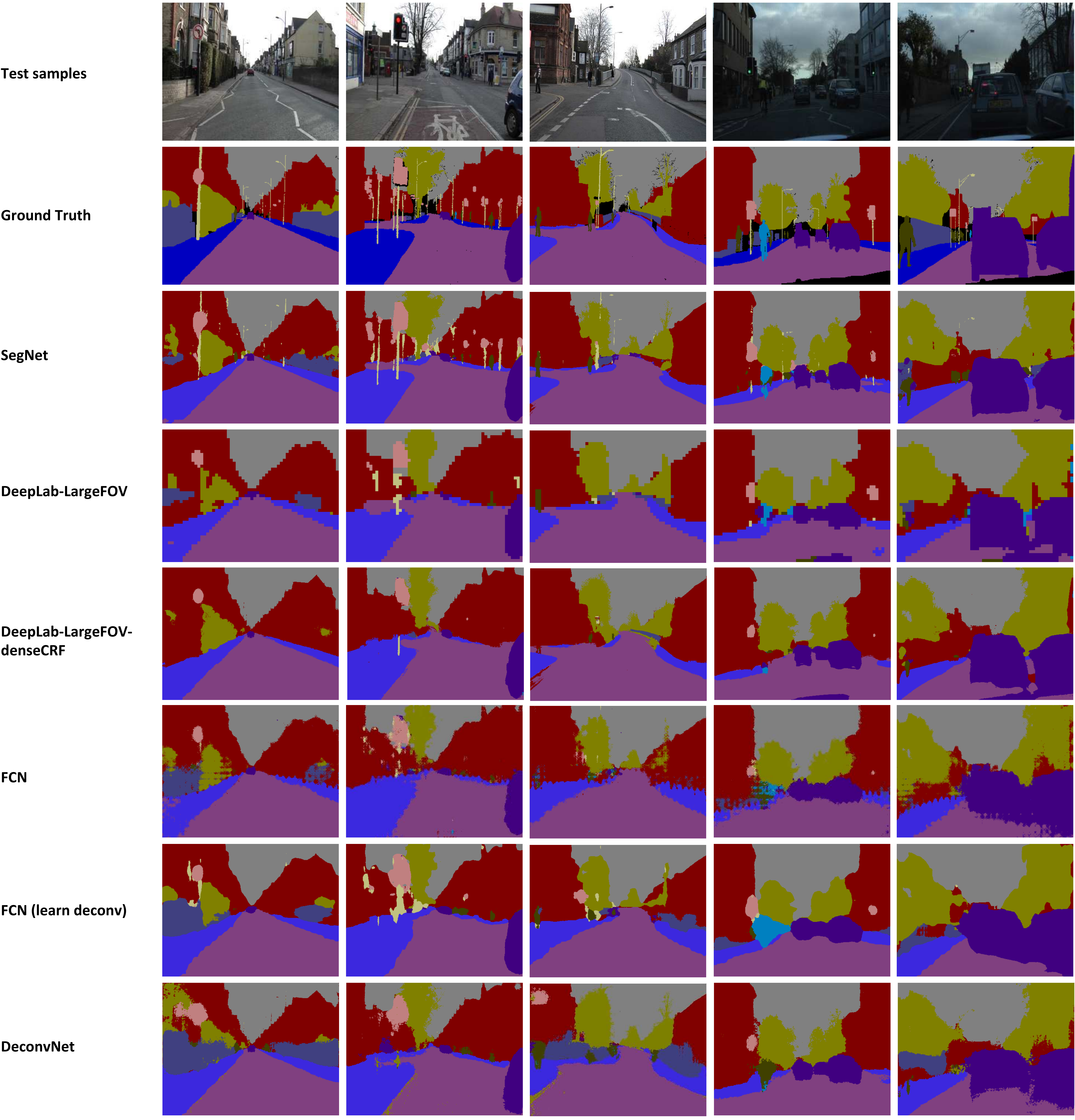}
	\caption{\footnotesize{Results on CamVid day and dusk test samples.
			SegNet shows superior performance, particularly with its ability to delineate boundaries, as compared to some of the larger models when all are trained in a controlled setting. DeepLab-LargeFOV is the most efficient model and with CRF post-processing can produce competitive results although smaller classes are lost. FCN with learnt deconvolution is clearly better. DeconvNet is the largest model with the longest training time, but its predictions loose small classes. Note that these results correspond to the model corresponding to the highest mIoU accuracy in Table \ref{CamvidDeepBenchmark}.
	}}
	\label{CamVidQualy}
\end{figure*}

\subsection{Road Scene Segmentation}
\label{CamVid}
A number of road scene datasets are available for semantic parsing \cite{gould2009decomposing,russell2008labelme,GabeDataset,GeigerKITTI}. Of these we choose to benchmark SegNet using the CamVid dataset \cite{GabeDataset} as it contains video sequences. This enables us to compare our proposed architecture with those which use \textit{motion and structure} \cite{LadickyECCV,Sturgess,Brostow} and video segments \cite{tighe2013superparsing}. We also combine \cite{gould2009decomposing,russell2008labelme,GabeDataset,GeigerKITTI} to form an ensemble of 3433 images to train SegNet for an additional benchmark. For a web demo (see footnote \ref{webdemo}) of road scene segmentation, we include the CamVid test set to this larger dataset. 
Here, we would like to note that another recent and independent segmentation benchmark on road scenes has been performed for SegNet and the other competing architectures used in this paper \cite{cordts2016cityscapes}. However, the benchmark was not controlled, meaning that each architecture was trained with a separate recipe with varying input resolutions and sometimes with a validation set included. Therefore, we believe our more controlled benchmark can be used to complement their efforts.

\begin{table*}[th]
	\resizebox{\textwidth}{!}{
		\small{
			\begin{tabular}{c|c|c|c|c|c|c|c|c|c|c|c|cccc}
				
				\multicolumn{1}{c}{Method}                   & \multicolumn{1}{c}{\rotatebox{90}{Building}} & \multicolumn{1}{c}{\rotatebox{90}{Tree}} & \multicolumn{1}{c}{\rotatebox{90}{Sky}}  & \multicolumn{1}{c}{\rotatebox{90}{Car}}  & \multicolumn{1}{c}{\rotatebox{90}{Sign-Symbol}} & \multicolumn{1}{c}{\rotatebox{90}{Road}} & \multicolumn{1}{c}{\rotatebox{90}{Pedestrian}} & \multicolumn{1}{c}{\rotatebox{90}{Fence}} & \multicolumn{1}{c}{\rotatebox{90}{Column-Pole}} & \multicolumn{1}{c}{\rotatebox{90}{Side-walk}} & \multicolumn{1}{c}{\rotatebox{90}{Bicyclist}} & \multicolumn{1}{c}{\rotatebox{90}{Class avg.}} & \multicolumn{1}{c}{\rotatebox{90}{Global avg.}} & \multicolumn{1}{c}{\rotatebox{90}{mIoU}}&\multicolumn{1}{c}{\rotatebox{90}{BF}}\\ \hline \hline
				
				
				SfM+Appearance  \cite{Brostow}           & 46.2     & 61.9 & 89.7 & 68.6 & 42.9        & 89.5 & 53.6       & 46.6  & 0.7         & 60.5     & 22.5      & 53.0       & 69.1  & \multicolumn{2}{c}{n/a$^{*}$}      \\ \hline
				
				Boosting    \cite{Sturgess}              & 61.9     & 67.3 & 91.1 & 71.1 & 58.5        & 92.9 & 49.5       & 37.6  & 25.8        & 77.8     & 24.7      & 59.8       & 76.4  & \multicolumn{2}{c}{n/a$^{*}$}      \\ \hline
				
				Dense Depth Maps   \cite{zhang2010semantic}          & 85.3     & 57.3 & 95.4 & 69.2 & 46.5        & \textbf{98.5} & 23.8       & 44.3  & 22.0        & 38.1     & 28.7      & 55.4       & 82.1    & \multicolumn{2}{c}{n/a$^{*}$}    \\ \hline
				
				Structured Random Forests \cite{kontschieder2011structured}& \multicolumn{11}{c|}{n/a}                                                                          & 51.4       & 72.5    &  \multicolumn{2}{c}{n/a$^{*}$}    \\ \hline
				
				Neural Decision Forests \cite{BuloNeural}  & \multicolumn{11}{c|}{n/a}                                                                          & 56.1       & 82.1   & \multicolumn{2}{c}{n/a$^{*}$}    \\ \hline
				
				Local Label Descriptors  \cite{yang2012local}  & 80.7     & 61.5 & 88.8 & 16.4 & n/a         & 98.0 & 1.09       & 0.05  & 4.13        & 12.4     & 0.07      & 36.3       & 73.6    & \multicolumn{2}{c}{n/a$^{*}$}    \\ \hline
				
				Super Parsing   \cite{tighe2013superparsing}           & 87.0     & 67.1 & 96.9 & 62.7 & 30.1        & 95.9 & 14.7       & 17.9  & 1.7         & 70.0     & 19.4      & 51.2       & 83.3    & \multicolumn{2}{c}{n/a$^{*}$}     \\ \hline
				SegNet (3.5K dataset training - 140K)          & \textbf{89.6}     & \textbf{83.4}  & 96.1 & \textbf{87.7} & 52.7 & 96.4 & \textbf{62.2} & \textbf{53.45} & \textbf{32.1}  & \textbf{93.3} & \textbf{36.5}  & \textbf{71.20}      & \textbf{90.40}   &60.10    &  46.84 \\ \hline
				\multicolumn{15}{c}{CRF based approaches}                                                                                                       \\ \hline
				
				Boosting + pairwise CRF  \cite{Sturgess} & 70.7     & 70.8 & 94.7 & 74.4 & 55.9        & 94.1 & 45.7       & 37.2  & 13.0        & 79.3     & 23.1      & 59.9       & 79.8  & \multicolumn{2}{c}{n/a$^{*}$}       \\ \hline
				
				Boosting+Higher order \cite{Sturgess}    & 84.5     & 72.6 & \textbf{97.5} & 72.7 & 34.1        & 95.3 & 34.2       & 45.7  & 8.1         & 77.6     & 28.5      & 59.2       & 83.8    & \multicolumn{2}{c}{n/a$^{*}$}     \\ \hline
				
				Boosting+Detectors+CRF \cite{LadickyECCV}   & 81.5     & 76.6 & 96.2 & 78.7 & 40.2        & 93.9 & 43.0       & 47.6  & 14.3        & 81.5     & 33.9      & 62.5       & 83.8    & \multicolumn{2}{c}{n/a$^{*}$}     \\ \hline
			\end{tabular}
	}}
	\vspace*{0.1cm}
	\caption{\footnotesize{Quantitative comparisons of SegNet with traditional methods on the CamVid 11 road class segmentation problem \cite{GabeDataset}. SegNet outperforms all the other methods, including those using depth, video and/or CRF's on the majority of classes. In comparison with the CRF based methods SegNet predictions are more accurate in 8 out of the 11 classes. It also shows a good $\approx 10\%$ improvement in class average accuracy when trained on a large dataset of 3.5K images. Particularly noteworthy are the significant improvements in accuracy for the smaller/thinner classes. * Note that we could not access predictions for older methods for computing the mIoU, BF metrics.
	}}
	\label{CamVidQuant}
\end{table*}

The qualitative comparisons of SegNet predictions with other deep architectures can be seen in Fig. \ref{CamVidQualy}. The qualitative results show the ability of the proposed architecture to segment smaller classes in road scenes while producing a smooth segmentation of the overall scene. Indeed, under the controlled benchmark setting, 
SegNet shows superior performance as compared to some of the larger models. DeepLab-LargeFOV is the most efficient model and with CRF post-processing can produce competitive results although smaller classes are lost. FCN with learnt deconvolution is clearly better than with fixed bilinear upsampling. DeconvNet is the largest model and the most inefficient to train. Its predictions do not retain small classes.


We also use this benchmark to first compare SegNet with several non deep-learning methods including Random Forests \cite{Jamie2}, Boosting \cite{Jamie2,Sturgess} in combination with CRF based methods \cite{LadickyECCV}. This was done to give the user a perspective of the improvements in accuracy that has been achieved using deep networks compared to classical feature engineering based techniques. 

The results in Table \ref{CamVidQuant} show SegNet-Basic, SegNet obtain competitive results  when compared with methods which use  CRFs. This shows the ability of the deep architecture to extract meaningful features from the input image and map it to accurate and smooth class segment labels. The most interesting result here is the large  performance improvement in class average and mIOU metrics that is obtained when a large training dataset, obtained by combining \cite{gould2009decomposing,russell2008labelme,GabeDataset,GeigerKITTI}, is used to train SegNet. Correspondingly, the qualitative results of SegNet (see Fig. \ref{CamVidQualy}) are clearly superior to the rest of the methods. It is able to segment both small and large classes well. We remark here that we used median frequency class balancing \cite{eigen2014NIPS} in training SegNet-Basic and SegNet. In addition, there is an overall smooth quality of segmentation much like what is typically obtained with CRF post-processing. Although the fact that results improve with larger training sets is not surprising, the percentage improvement obtained using pre-trained encoder network and this training set indicates that this architecture can potentially be deployed for practical applications. Our random testing on urban and highway images from the internet (see Fig. \ref{Teaser}) demonstrates that SegNet can \textit{absorb} a large training set and generalize well to unseen images. It also indicates the contribution of the prior (CRF) can be lessened when sufficient amount of training data is made available.

In Table \ref{CamvidDeepBenchmark} we compare SegNet's performance with now widely adopted fully convolutional architectures for segmentation. As compared to the experiment in Table \ref{CamVidQuant}, we did not use any class blancing for training any of the deep architectures including SegNet. This is because we found it difficult to train larger models such as DeconvNet with median frequency balancing. We benchmark performance at 40K, 80K and $>$80K iterations which given the mini-batch size and training set size approximately corresponds to $50,100$ and $>$100 epochs. For the last test point we also report the maximum number of iterations (here atleast 150 epochs) beyond which we observed no accuracy improvements or when over-fitting set in. We report the metrics at three stages in the training phase to reveal how the metrics varied with training time, particularly for larger networks. This is important to understand if additional training time is justified when set against accuracy increases. Note also that for each evaluation we performed a complete run through the dataset to obtain batch norm statistics and then evaluated the test model with this statistic (see \url{http://mi.eng.cam.ac.uk/projects/segnet/tutorial.html} for code.). These evaluations are expensive to perform on large training sets and hence we only report metrics at three time points in the training phase. 

From Table \ref{CamvidDeepBenchmark} we immediately see that SegNet, DeconvNet achieve the highest scores in all the metrics as compared to other models. DeconvNet has a higher boundary delineation accuracy but  SegNet is much more efficient as compared to DeconvNet. This can be seen from the compute statistics in Table \ref{TimeBenchmark}. FCN, DeconvNet which have  fully connected layers (turned into convolutional layers) train much more slowly and have comparable or higher forward-backward pass time with reference to SegNet. Here we note also that over-fitting was not an issue in training these larger models, since at comparable iterations to SegNet their metrics showed an increasing trend. 

For the FCN model learning the deconvolutional layers as opposed to fixing them with bi-linear interpolation weights improves performance particularly the BF score. It also achieves higher metrics in a far lesser time. This fact agrees with our earlier analysis in Sec. \ref{Analysis}. 

Surprisingly, DeepLab-LargeFOV which is trained to predict labels at a resolution of $45\times60$ produces competitive performance given that it is the smallest model in terms of parameterization and also has the fastest training time as per Table \ref{TimeBenchmark}. However, the boundary accuracy is poorer and this is shared by the other architectures. DeconvNet's BF score is higher than the other networks when trained for a very long time. Given our analysis in Sec. \ref{Analysis} and the fact that it shares a SegNet type architecture. 

The impact of dense CRF \cite{koltun2011efficient} post-processing can be seen in the last time point for DeepLab-LargeFOV-denseCRF. Both global and mIoU improve but class average diminshes. However a large improvement is obtained for the BF score. Note here that the dense CRF hyperparameters were obtained by an expensive grid-search process on a subset of the training set since no validation set was available.

\begin{table*}[t]
\centering
\tabcolsep=2pt
\begin{tabular}{|c|c|c|c|c||c|c|c|c||c|c|c|c|c|}
\hline
Network/Iterations       & \multicolumn{4}{c||}{40K} & \multicolumn{4}{c||}{80K} & \multicolumn{4}{c|}{$>$80K} & Max iter \\ \hline \hline
  & G & C & mIoU & BF & G & C & mIoU & BF & G & C & mIoU & BF & \\ \hline 

SegNet   & 88.81 & 59.93 & 50.02 & 35.78 & 89.68 & 69.82 &   57.18 & 42.08 &90.40 & 71.20  &60.10  & 46.84 & 140K\\ \hline
DeepLab-LargeFOV\cite{liang2015semantic} & 85.95 & 60.41 & 50.18 & 26.25 & 87.76 & 62.57 & 53.34 & 32.04 & 88.20 & 62.53 & 53.88 & 32.77 & 140K\\ \hline
DeepLab-LargeFOV-denseCRF\cite{liang2015semantic} & \multicolumn{8}{c|}{not computed} & 89.71 & 60.67 & 54.74 & 40.79 & 140K\\ \hline
FCN & 81.97 & 54.38 & 46.59 & 22.86  &82.71 & 56.22 & 47.95 & 24.76& 83.27 & 59.56 & 49.83 & 27.99 & 200K \\  \hline
FCN (learnt deconv) \cite{FCN} &83.21 &56.05 & 48.68 &27.40 & 83.71 & 59.64 & 50.80 & 31.01 & 83.14 & 64.21 & 51.96 & 33.18 & 160K \\ \hline
DeconvNet \cite{noh2015learning} & 85.26 & 46.40 & 39.69 & 27.36 & 85.19 & 54.08 &43.74 & 29.33 & 89.58 & 70.24 & 59.77 & 52.23 & 260K \\ \hline 
\end{tabular}
\vspace*{0.1cm}
\caption{\footnotesize{Quantitative comparison of deep networks for semantic segmentation on the CamVid test set when trained on a corpus of 3433 road scenes \textit{without class balancing}. When end-to-end training is performed with the same and fixed learning rate, smaller networks like SegNet learn to perform better in a shorter time. The BF score which measures the accuracy of inter-class boundary delineation is significantly higher for SegNet, DeconvNet as compared to other competing models. DeconvNet matches the metrics for SegNet but at a much larger computational cost.  Also see Table \ref{CamVidQuant} for individual class accuracies for SegNet.}}
\label{CamvidDeepBenchmark}
\end{table*}

\begin{table*}[t]
	\centering
	\tabcolsep=2pt
	\begin{tabular}{|c|c|c|c|c||c|c|c|c||c|c|c|c|c|}
		\hline
		Network/Iterations       & \multicolumn{4}{c||}{80K} & \multicolumn{4}{c||}{140K} & \multicolumn{4}{c|}{$>$140K} & Max iter  \\ \hline \hline
		& G & C & mIoU & BF & G & C & mIoU & BF & G & C & mIoU & BF  &\\ \hline 
		
		SegNet   &70.73  & 30.82 & 22.52 & 9.16  &71.66  &37.60  &27.46  & 11.33 & 72.63 & 44.76 & 31.84 & 12.66 & 240K\\ \hline
		DeepLab-LargeFOV \cite{liang2015semantic}&70.70 &41.75 &30.67 &7.28 &71.16 & 42.71 & 31.29 & 7.57 & 71.90 & 42.21 & 32.08 & 8.26 & 240K\\ \hline
		DeepLab-LargeFOV-denseCRF \cite{liang2015semantic}& \multicolumn{8}{c|}{not computed} & 66.96 & 33.06 & 24.13& 9.41 & 240K \\ \hline
		FCN (learnt deconv) \cite{FCN}& 67.31 & 34.32 & 24.05 & 7.88 & 68.04 & 37.2 & 26.33 & 9.0 & 68.18 & 38.41 & 27.39 & 9.68 & 200K \\  \hline
		DeconvNet \cite{noh2015learning}& 59.62 & 12.93 & 8.35 & 6.50 & 63.28 & 22.53 & 15.14 & 7.86 & 66.13 & 32.28 & 22.57 & 10.47 & 380K \\ \hline 
	\end{tabular}
	\vspace*{0.1cm}
	\caption{\footnotesize{Quantitative comparison of deep architectures on the SUNRGB-D dataset when trained on a corpus of 5250 indoor scenes. Note that only the RGB modality was used in these experiments. In this complex task with $37$ classes all the architectures perform poorly, particularly because of the smaller sized classes and skew in the class distribution. DeepLab-Large FOV, the smallest and most efficient model has a slightly higher mIoU but SegNet has a better G,C,BF score. Also note that when SegNet was trained with \textit{median frequency class balancing} it obtained 71.75, 44.85, 32.08, 14.06 (180K) as the metrics.}}
	\label{SUNRGBBenchmark}
\end{table*}

\subsection{SUN RGB-D Indoor Scenes}
\label{SUNRGBD}
SUN RGB-D \cite{song2015sun} is a very challenging and large dataset of indoor scenes with $5285$ training and $5050$ testing images. The images are captured by different sensors and hence come in various resolutions. The task is to segment $37$ indoor scene classes including wall, floor, ceiling, table, chair, sofa etc. This task is made hard by the fact that object classes come in various shapes, sizes and in different poses. There are frequent partial occlusions since there are typically many different classes present in each of the test images. These factors make this one of the hardest segmentation challenges. We only use the RGB modality for our training and testing. Using the depth modality would necessitate architectural modifications/redesign \cite{FCN}. Also the quality of depth images from current cameras require careful post-processing to fill-in missing measurements. They may also require using fusion of many frames to robustly extract features for segmentation. Therefore we believe using depth for segmentation merits a separate body of work which is not in the scope of this paper. We also note that an earlier benchmark dataset NYUv2 \cite{silberman2012indoor} is included as part of this dataset.

\begin{figure*}
\centering
\includegraphics[width=0.9\textwidth]{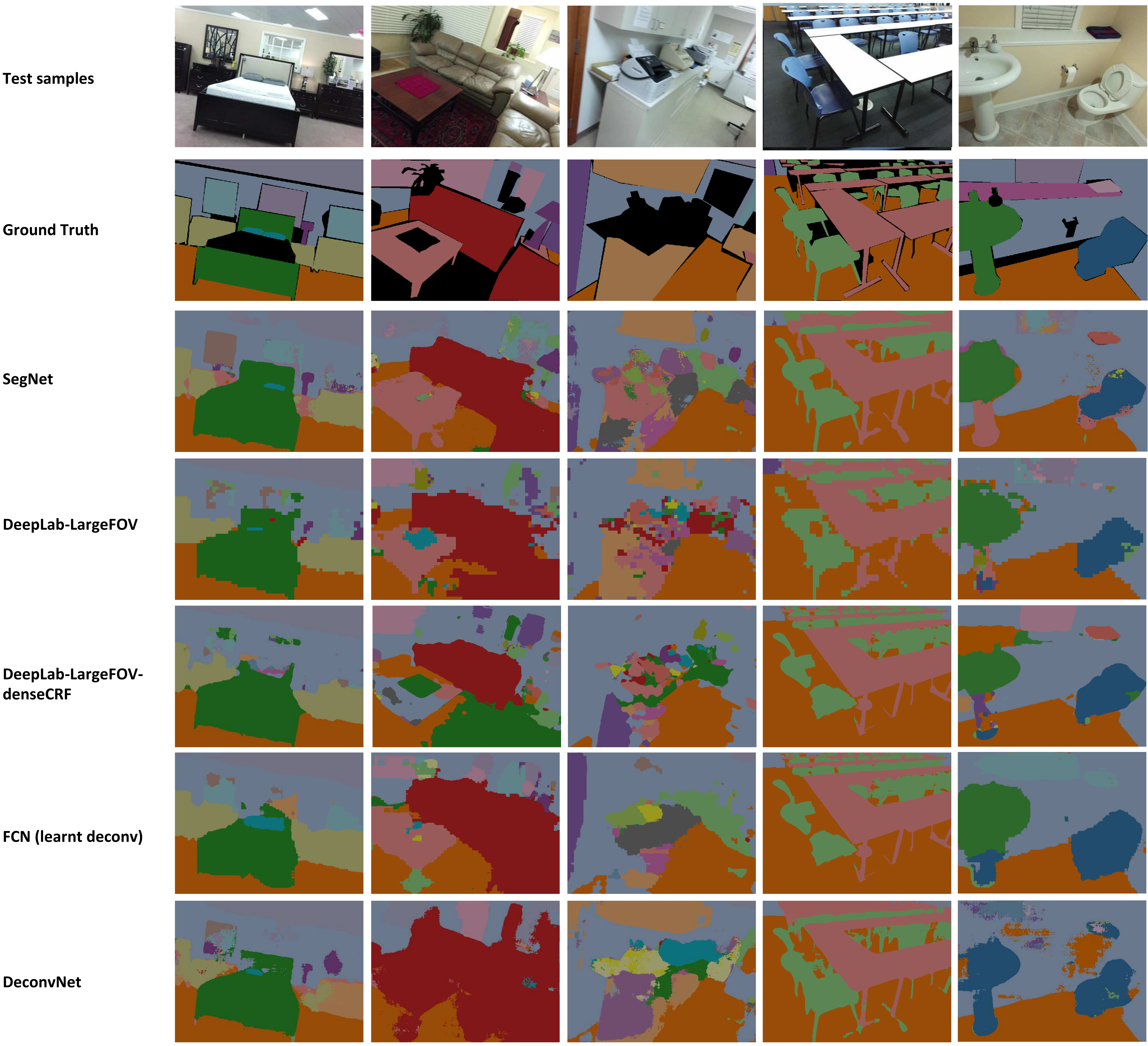}
\caption{\footnotesize{Qualitative assessment of SegNet predictions on RGB indoor test scenes from the recently released SUN RGB-D dataset \cite{song2015sun}. In this hard challenge, SegNet predictions delineate inter class boundaries well for object classes in a variety of scenes and their view-points. Overall rhe segmentation quality is better when object classes are reasonably sized but is very noisy when the scene is more cluttered. Note that often parts of an image of a scene do not have ground truth labels and these are shown in black colour. These parts are not masked in the corresponding deep model  predictions that are shown. Note that these results correspond to the model corresponding to the highest mIoU accuracy in Table \ref{SUNRGBBenchmark}.}}
\label{SUNRGBDQualy}
\end{figure*}

Road scene images have limited variation, both in terms of the classes of interest and their spatial arrangements. When captured from a moving vehicle where the camera position is nearly always parallel to the road surface limiting variability in view points. This makes it easier for deep networks to learn to segment them robustly. In comparison, images of indoor scenes are more complex since the view points can vary a lot and there is less regularity in both the number of classes present in a scene and their spatial arrangement. Another difficulty is caused by the widely varying sizes of the object classes in the scene. Some test samples from the recent SUN RGB-D dataset \cite{song2015sun} are shown in Fig. \ref{SUNRGBDQualy}. We observe some scenes with few large classes and some others with dense clutter (bottom row and right). The appearance (texture and shape) can also widely vary in indoor scenes. Therefore, we believe this is the hardest challenge for segmentation architectures and methods in computer vision. Other challenges, such as Pascal VOC12 \cite{Pascal} salient object segmentation have occupied researchers  more \cite{liu2015semantic}, but we believe indoor scene segmentation is more challenging and has more current practical applications such as in AR and robotics. To encourage more research in this direction we compared well known deep architectures on the large SUN RGB-D dataset.



The qualitative results of SegNet on samples of indoor scenes of different types such as bedroom, living room, laboratory, meeting room, bathroom are shown in Fig. \ref{SUNRGBDQualy}. We see that SegNet obtains reasonable predictions when the size of the classes are large under different view points. This is particularly interesting since the input modality is only RGB. 
RGB images are also useful to segment thinner structures such as the legs of chairs and tables, lamps which is difficult to achieve using depth images from currently available sensors. This can be seen from the results of SegNet, DeconvNet in Fig. \ref{SUNRGBDQualy}. It is also useful to segment decorative objects such as paintings on the wall for AR tasks. However as compared to outdoor scenes the segmentation quality is clearly more noisy. The quality drops significantly when clutter is increased (see the result sample in the middle column).

The quantitative results in Table \ref{SUNRGBBenchmark} show that all the deep architectures share low mIoU and boundary metrics. The global and class averages (correlates well with mIou) are also small. SegNet outperforms all other methods in terms of G,C, BF metrics and has a slightly lower mIoU than DeepLab-LargeFOV. As a stand alone experiment we trained SegNet with median frequency class balancing \cite{eigen2014predicting} and the metrics were higher (see Table \ref{SUNRGBBenchmark}) and this agrees with our analysis in Sec. \ref{Analysis}. Interestingly, using the grid search based optimal hyperparameters for the dense-CRF worsened all except the BF score metric for DeepLab-LargeFOV-denseCRF. More optimal settings could perhaps be found but the grid search process was too expensive given the large inference time for dense-CRFs. 

One reason for the overall poor performance is the large number of classes in this segmentation task, many of which occupy a small part of the image and appear infrequently. The accuracies reported in Table \ref{SUNRGBDClassavg} clearly show that larger classes have reasonable accuracy and smaller classes have lower accuracies. This can be improved with larger sized datasets and class distribution aware training techniques. Another reason for poor performance could lie in the inability of these deep architectures (all are based on the VGG architecture \cite{Simonyan}) to large variability in indoor scenes . This conjecture on our part is based on the fact that the smallest model DeepLab-LargeFOV produces the best accuracy in terms of mIoU and in comparison, larger parameterizations in DeconvNet, FCN did not improve perfomance even with much longer training (DeconvNet). This suggests there could lie a common reason for poor performance across all architectures. More controlled datasets \cite{handa2015scenenet} are needed to verify this hypothesis.

\section{Discussion and future work}
\label{Discussion}
Deep learning models have often achieved increasing success due to the availability of massive datasets and expanding model depth and parameterisation. However, in practice factors like memory and computational time during training and testing are important factors to consider when choosing a model from a large bank of models. 
Training time becomes an important consideration particularly when the performance gain is not commensurate with increased training time as shown in our experiments. Test time memory and computational load are important to deploy models on specialised embedded devices, for example, in AR applications. 
From an overall efficiency viewpoint, we feel less attention has been paid to smaller and more memory, time efficient models for real-time applications such as road scene understanding and AR. This was the primary motivation behind the proposal of SegNet, which is significantly smaller and faster than other competing architectures, but which we have shown to be efficient for tasks such as road scene understanding. 
 
Segmentation challenges such as Pascal \cite{Pascal} and MS-COCO \cite{COCO} are object segmentation challenges wherein a few classes are present in any test image.  Scene segmentation is more challenging due to the high variability of indoor scenes and a need to segment a larger number of classes simultaneously. The task of outdoor and indoor scene segmentation are also more practically oriented with current applications such as autonomous driving, robotics and AR. 

 The metrics we chose to benchmark various deep segmentation architectures like the boundary F1-measure (BF) was done to complement the existing metrics which are more biased towards region accuracies. It is clear from our experiments and other independent benchmarks \cite{cordts2016cityscapes} that outdoor scene images captured from a moving car are easier to segment and deep architectures perform robustly. We hope our experiments will encourage researchers to engage their attention towards the more challenging indoor scene segmentation task.
 
An important choice we had to make when benchmarking different deep architectures of varying parameterization was the manner in which to train them. Many of these architectures have used a host of supporting techniques and multi-stage training recipes to arrive at high accuracies on datasets but this makes it difficult to gather evidence about their true performance under time and memory constraints. Instead we chose to perform a controlled benchmarking where we used batch normalization to enable end-to-end training with the same solver (SGD). However, we note that this approach cannot entirely disentangle the effects of model versus solver  (optimization) in achieving a particular result. This is mainly due to the fact that training these networks involves gradient back-propagation which is imperfect and the optimization is a non-convex problem in extremely large dimensions. Acknowledging these shortcomings, our hope is that this controlled analysis complements other benchmarks \cite{cordts2016cityscapes} and reveals the practical trade-offs involved in different well known architectures.

For the future, we would like to exploit our understanding of segmentation architectures gathered from our analysis to design more efficient architectures for real-time applications. We are also interested in estimating the model uncertainty for predictions from deep segmentation architectures \cite{gal2015dropout},  \cite{kendall2015bayesian}.

\begin{table*}[t]
	\centering
	\tabcolsep=3pt
	\begin{tabular}{c|c|c|c|c|c|c|c|c|c|c|c|c}
		\multicolumn{1}{c|}{Wall} & \multicolumn{1}{c|}{Floor} & \multicolumn{1}{c|}{Cabinet} & \multicolumn{1}{c|}{Bed} & \multicolumn{1}{c|}{Chair} & \multicolumn{1}{c|}{Sofa} & \multicolumn{1}{c|}{Table} & \multicolumn{1}{c|}{Door} & \multicolumn{1}{c|}{Window} & \multicolumn{1}{c|}{Bookshelf} & \multicolumn{1}{c|}{Picture} & \multicolumn{1}{c|}{Counter} & \multicolumn{1}{c}{Blinds} \\ \hline
		83.42                    & 93.43                       & 63.37                        & 73.18                     & 75.92                      & 59.57                      &       64.18                 & 52.50                      & 57.51                        & 42.05                           & 56.17                         & 37.66                        & 40.29                      \\ \hline
		Desk                       & Shelves                    & Curtain                      & Dresser                  & Pillow                     & Mirror                    & Floor mat                  & Clothes                   & Ceiling                     & Books                          & Fridge                       & TV                           & Paper                       \\ \hline
		11.92                      & 11.45                        & 66.56                         & 52.73                     & 43.80                       & 26.30                     & 0.00                       & 34.31                      & 74.11                        & 53.77                           & 29.85                         & 33.76                         & 22.73                       \\ \hline
		Towel                      & Shower curtain             & Box                          & Whiteboard               & Person                     & Night stand               & Toilet                     & Sink                      & Lamp                        & Bathtub                        & Bag                          & \multicolumn{2}{c}{\multirow{2}{*}{}}                     \\ \cline{1-11}
		19.83                       & 0.03                        & 23.14                          & 60.25                     & 27.27                       & 29.88                     & 76.00                       & 58.10                     & 35.27                        & 48.86                           & 16.76                          & \multicolumn{2}{c}{}                                      \\ \cline{1-11}
	\end{tabular}
	\vspace*{0.1cm}
	\caption{Class average accuracies of SegNet predictions for the 37 indoor scene classes in the SUN RGB-D benchmark dataset. The performance correlates well with size of the classes in indoor scenes. Note that class average accuracy has a strong correlation with mIoU metric.}
	\label{SUNRGBDClassavg}
\end{table*}

\begin{table*}[t]
	\centering
	\tabcolsep=2pt
	\begin{tabular}{|c|c|c|c|c|c|}
		\hline
		Network       & Forward pass(ms) & Backward pass(ms) & GPU training memory (MB) & GPU inference memory (MB) & Model size (MB)  \\ \hline \hline
		SegNet  & 422.50 & 488.71 & 6803 & \textbf{1052} & 117\\ \hline
		DeepLab-LargeFOV \cite{liang2015semantic} & \textbf{110.06} & \textbf{160.73}& \textbf{5618} & 1993 & 83 \\ \hline
		FCN (learnt deconv)\cite{FCN} & 317.09 & 484.11 & 9735 & 1806 & 539 \\ \hline
		DeconvNet \cite{noh2015learning} & 474.65 & 602.15 & 9731 & 1872 & 877 \\ \hline
	\end{tabular}
	\vspace*{0.1cm}
	\caption{\footnotesize{A comparison of computational time and hardware resources required for various deep architectures. The caffe time command was used to compute time requirement averaged over 10 iterations with mini batch size 1 and an image of $360\times480$ resolution We used nvidia-smi unix command to compute memory consumption. For training memory computation we used a mini-batch of size 4 and for inference memory the batch size was 1. Model size was the size of the caffe models on disk. SegNet is most memory efficient during inference model.}}
	\label{TimeBenchmark}
\end{table*}

\section{Conclusion}
\label{Conclusion}
We presented SegNet, a deep convolutional network architecture for semantic segmentation. The main motivation behind SegNet was the need to design an efficient architecture for road and indoor scene understanding which is efficient both in terms of memory and computational time. We analysed SegNet and compared it with other important variants to reveal the practical trade-offs involved in designing architectures for segmentation, particularly training time, memory versus accuracy. Those architectures which store the encoder network feature maps in full perform best but consume more memory during inference time. SegNet on the other hand is more efficient since it only stores the max-pooling indices of the feature maps and uses them in its decoder network to achieve good performance. On large and well known datasets SegNet performs competitively, achieving high scores for road scene understanding. End-to-end learning of deep segmentation architectures is a harder challenge and we hope to see more attention paid to this important problem.

\ifCLASSOPTIONcaptionsoff
  \newpage
\fi

\bibliographystyle{ieeetr}
\bibliography{RefBase}


\begin{IEEEbiography}[{\includegraphics[width=1in,height=1.25in,clip,keepaspectratio]{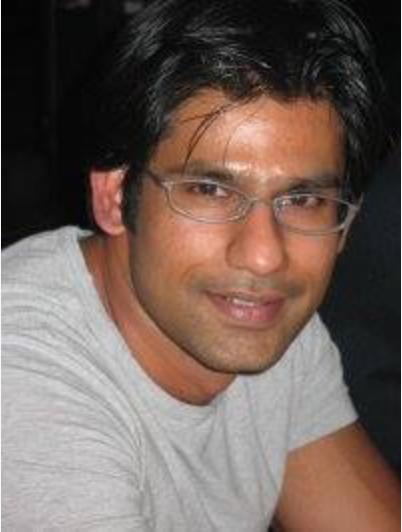}}]{Vijay Badrinarayanan} obtained his Ph.D from INRIA Rennes, France in 2009. He was a senior post-doctoral research associate at the Machine Intelligence Laboratory, Department of Engineering, University of Cambridge, U.K. He currently works as a Principal Engineer, Deep Learning at Magic Leap, Inc. in Mountain View, CA. His research interests are in probabilistic graphical models, deep learning applied to image and video based perception problems.
\end{IEEEbiography}

\vspace{-1cm}

\begin{IEEEbiography}[{\includegraphics[width=1in,height=1.25in,clip,keepaspectratio]{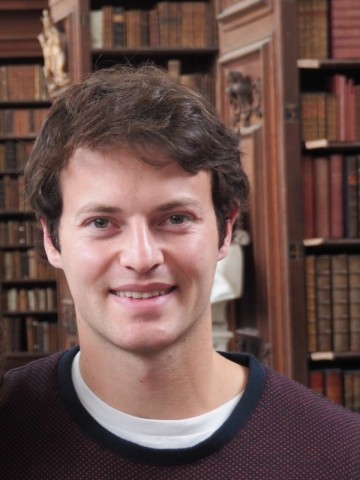}}]{Alex Kendall} graduated with a Bachelor of Engineering with First Class Honours in 2013 from the University of Auckland, New Zealand. In 2014 he was awarded a Woolf Fisher Scholarship to study towards a Ph.D at the University of Cambridge, U.K. He is a member of the Machine Intelligence Laboratory and is interested in applications of deep learning for mobile robotics.
\end{IEEEbiography}

\vspace{-1cm}

\begin{IEEEbiography}[{\includegraphics[width=1in,height=1.25in,clip,keepaspectratio]{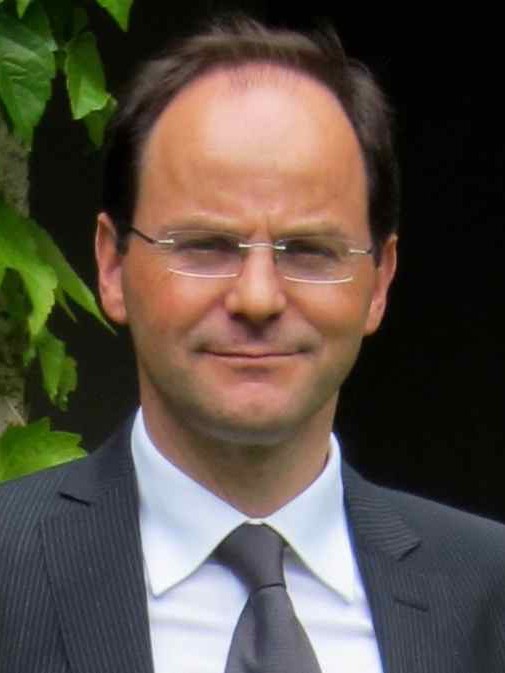}}]{Roberto Cipolla} obtained a B.A. (Engineering) degree from the University of Cambridge in 1984, an M.S.E. (Electrical Engineering) from the University of Pennsylvania in 1985 and a D.Phil. (Computer Vision) from the University of Oxford in 1991. from 1991-92 was a Toshiba Fellow and engineer at the Toshiba Corporation Research and Development Centre in Kawasaki, Japan. He joined the Department of Engineering, University of Cambridge in 1992 as a Lecturer and a Fellow of Jesus College. He became a Reader in Information Engineering in 1997 and a Professor in 2000. He became a Fellow of the Royal Academy of Engineering (FREng) in 2010. His research interests are in computer vision and robotics. He has authored 3 books, edited 9 volumes and co-authored more than 300 papers.
\end{IEEEbiography}

\end{document}